\documentclass[11pt,a4paper]{article}
\usepackage[hyperref]{acl2018}
\usepackage{acl2018}
\usepackage{times}
\usepackage{latexsym}
\usepackage{graphicx}
\usepackage{footnote}
\usepackage{amsmath,cleveref }
\usepackage[normalem]{ulem}
\usepackage{amsfonts}
\makeatother
\usepackage{url}

\aclfinalcopy 
\def\aclpaperid{1024} 

\usepackage{algorithm2e}
\usepackage{listings}

\DeclareMathOperator*{\argmax}{argmax}
\def\BState{\State\hskip-\ALG@thistlm}
\usepackage{centernot}

\title{AMR Parsing as Graph Prediction with Latent Alignment}

\author{
Chunchuan Lyu$^1$ ~~  Ivan Titov$^{1,2}$ \\
$^1$ILCC, School of Informatics, University of Edinburgh \\
$^2$ILLC, University of Amsterdam \\ 
}

\date{}
\begin{document}
\maketitle
\begin{abstract}
Abstract meaning representations (AMRs) are broad-coverage sentence-level semantic representations. AMRs represent sentences as rooted labeled directed acyclic graphs. AMR parsing is challenging partly due to the lack of annotated alignments between
nodes in the graphs and words in the corresponding sentences.
We introduce a neural parser which treats alignments as latent variables within a joint probabilistic model of concepts, relations and alignments. 
As exact inference requires marginalizing over alignments and 
is infeasible, we use the variational auto-encoding framework and a continuous relaxation of the discrete alignments.
We show that joint modeling 
is preferable to using a pipeline of align and parse. 
The parser achieves the best reported results on the standard benchmark (74.4\% on LDC2016E25).

\end{abstract}

\section{Introduction}

Abstract meaning representations (AMRs) \cite{Banarescu13abstractmeaning} are broad-coverage sentence-level semantic representations. 
AMR encodes, among others, information about semantic relations, named entities, co-reference, negation and modality.
\begin{figure}[ht!]
\centering
\includegraphics[width=0.8\columnwidth]{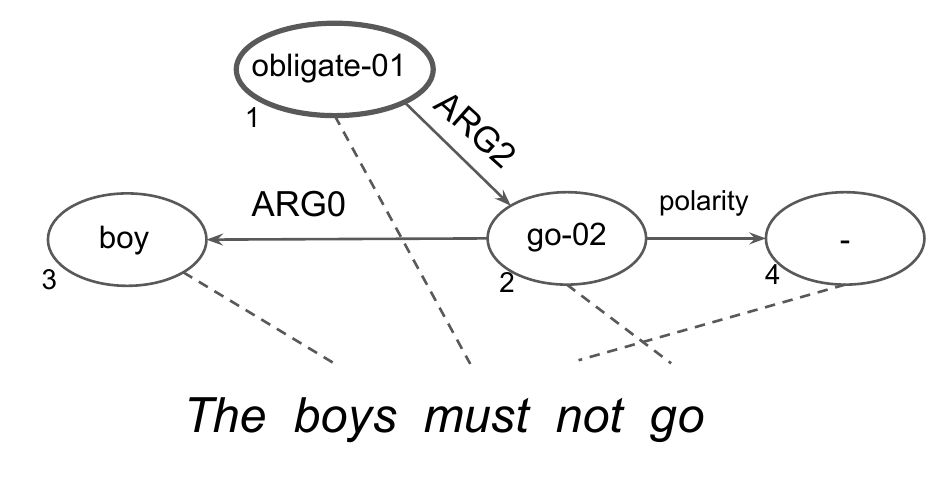}
\vspace{-2ex}
\caption{An example of AMR, the dashed lines denote latent alignments, {\it obligate-01} is the root. Numbers indicate depth-first traversal order.}
\label{fig:amr-example}
\end{figure}
The semantic representations can be regarded as rooted labeled directed acyclic graphs (see Figure ~\ref{fig:amr-example}). As AMR abstracts away from details of surface realization, it is potentially beneficial in many semantic related NLP tasks, including text summarization~\cite{Liu2015TowardAS,dohare2017text}, machine translation~\cite{Jones2012SemanticsBasedMT} and question answering~\cite{AAAI1612345}. 

AMR parsing has recently received a lot of attention~(e.g., \cite{Flanigan_adiscriminative,Artzi2015BroadcoverageCS,konstas-EtAl:2017:Long}). One distinctive aspect of AMR annotation is the lack of explicit alignments between nodes in the graph ({\it concepts}) and words in the sentences. Though this arguably simplified the annotation process~\cite{Banarescu13abstractmeaning}, it is not straightforward to produce an effective parser without relying on an alignment. 
Most AMR parsers~\cite{Marco,jamr-16,werling2015robust,wang2017getting,foland-martin:2017:Long}
use a pipeline where the aligner training stage precedes training a parser.
The aligners are not directly informed by the AMR parsing objective and may produce alignments suboptimal for this task.

In this work, we demonstrate that the alignments can be treated as latent variables in a joint probabilistic model and induced in such a way as to be beneficial for AMR parsing.
Intuitively, in our probabilistic model, every node in a graph is assumed to be aligned to a word in a sentence: each concept is predicted based on the corresponding RNN state.
Similarly, graph edges (i.e. relations) are predicted based on representations of concepts and aligned words (see Figure~\ref{fig:rel-pred}). As alignments are latent, exact inference requires marginalizing over latent alignments, which is infeasible. Instead we use variational inference, specifically the variational autoencoding framework of~\newcite{kingma2013auto}. Using discrete latent variables in deep learning has proven to be challenging~\cite{mnih2014neural,bornschein2015reweighted}.  We use a continuous relaxation of the alignment problem, relying on the recently introduced Gumbel-Sinkhorn construction~\cite{sinkhorn}.
This yields a computationally-efficient approximate method for estimating our joint probabilistic model of concepts, relations and alignments.

We assume injective alignments from concepts to words: every node in the graph is aligned to a single word in the sentence and every word is aligned to at most one node in the graph. 
This is necessary for two reasons. First, it lets us treat concept identification as sequence tagging at test time. For every word we would simply predict the corresponding concept or predict {\it NULL} to signify that no concept should be generated at this position.
Secondly, Gumbel-Sinkhorn can only work under this 
assumption. This constraint, though often appropriate, is problematic for certain AMR constructions  (e.g., named entities). In order to deal with these cases, we re-categorized AMR concepts. Similar recategorization strategies have been used in previous work~\cite{foland-martin:2017:Long,peng2017addressing}.

The resulting parser achieves 74.4\% Smatch score on the standard test set when using LDC2016E25 training set,\footnote{The standard deviation across multiple training runs was 0.16\%.} 
an improvement of 3.4\% over the previous best result~\cite{Character}. 
We also demonstrate that inducing alignments within the joint model is indeed beneficial. When, instead  of inducing alignments, 
we follow the standard approach and produce them on preprocessing,  the performance drops by 0.9\% Smatch.
Our main contributions can be summarized as
follows:
\begin{itemize}
\item we introduce a joint probabilistic model for alignment, concept and relation identification;
\item we demonstrate that a continuous relaxation can be used to effectively estimate the model;
\item the model achieves the best reported results.\footnote{The code can be accessed from \url{https://github.com/ChunchuanLv/AMR_AS_GRAPH_PREDICTION}}
\end{itemize}

\section{Probabilistic Model}

In this section we describe our probabilistic model and the estimation technique. 
In section~\ref{sect:prepost}, we describe preprocessing and post-processing (including concept re-categorization,  sense disambiguation, wikification  and root selection).

\subsection{Notation and setting}

We will use the following notation throughout the paper.
We refer to words in the sentences as $\mathbf{w} = (w_1,\ldots,w_n)$, where $n$ is sentence length, $w_k \in \mathcal{V}$ for $k \in \{1\ldots,n\}$. The concepts (i.e. labeled nodes) are $\mathbf{c} = (c_1, \ldots, c_m)$, where $m$ is the number of concepts and $c_i \in \mathcal{C}$ for $i \in \{1\ldots,m\}$. For example, in Figure~\ref{fig:amr-example}, $\mathbf{c} = (\mbox{\it obligate}, \mbox{\it go}, \mbox{\it boy}, \mbox{\it -})$.\footnote{The probabilistic model is invariant to the ordering of concepts, though the order affects the inference algorithm (see Section \ref{sec:align-mod}). We use depth-first traversal of the graph to generate the ordering.} Note that senses are predicted at post-processing, as discussed in Section~\ref{sec:post} (i.e. {\it go} is labeled as {\it go-02}).

A relation between `predicate concept' $i$
and `argument concept' $j$ is denoted by $r_{ij} \in \mathcal{R}$; it is set to \textit{NULL} if $j$ is not an argument of $i$.
In our example, $r_{2,3} = \mbox{\it ARG0}$ and $r_{1,3} = \mbox{\it NULL}$. We will use $R$ to denote all relations in the graph. 

To represent alignments, we will use
$\mathbf{a} = \{a_1, \ldots, a_m\}$, where $a_i\in \{1, \ldots, n\}$ returns the index of a word aligned to concept~$i$. In our example, $a_{1} = 3$.

All three model components rely on  bi-directional LSTM encoders~\cite{BiLSTM}. We denote states of BiLSTM (i.e. concatenation of forward and backward LSTM states) as $\mathbf{h}_k \in \mathbb{R}^d$ ($k \in \{1,\ldots,n\}$).  The sentence encoder takes pre-trained fixed word embeddings, randomly initialized lemma embeddings, part-of-speech  and named-entity tag embeddings.

\subsection{Method overview}
\label{sec:overview}

\begin{figure}[h!]
\centering
\includegraphics[width=\columnwidth]{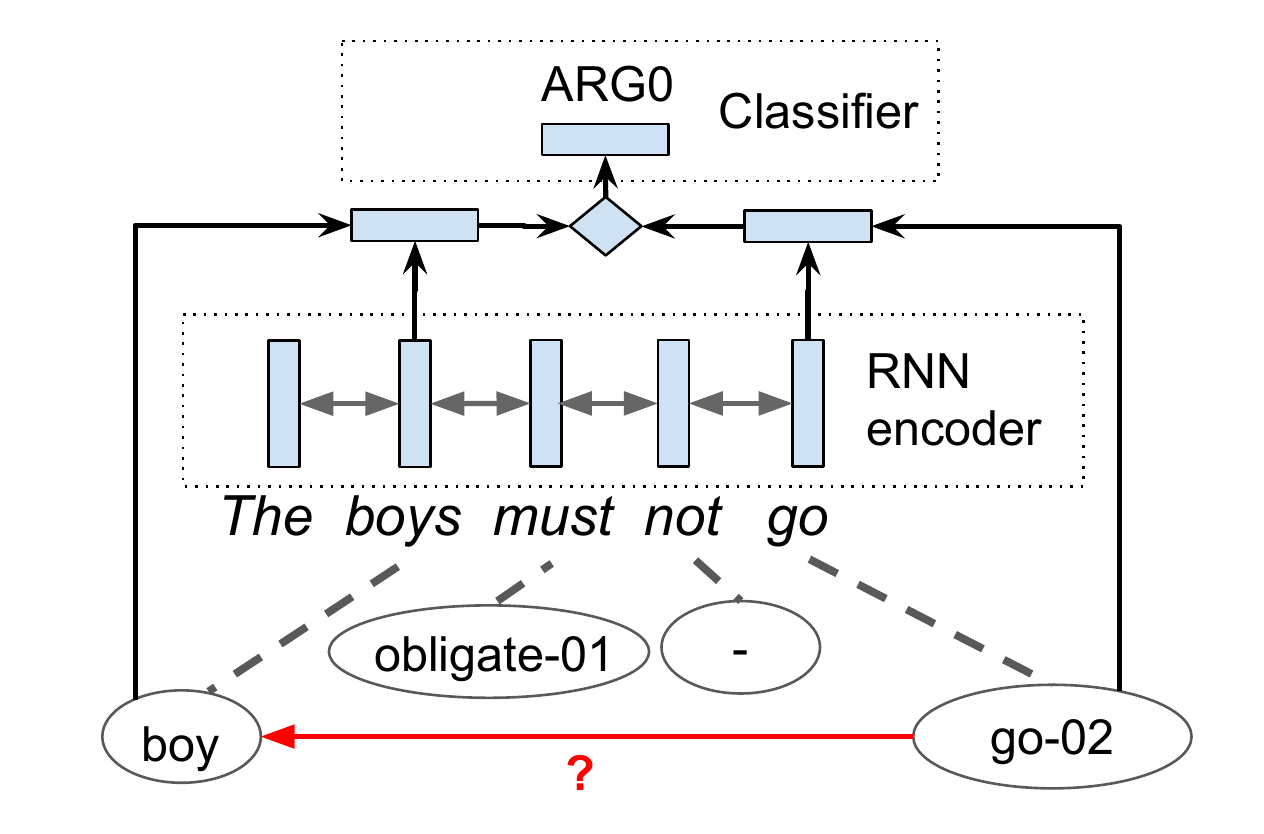}
\vspace{-3ex}
\caption{Relation identification: predicting a relation between  {\it boy} and {\it go-02} relying on 
the two concepts and corresponding RNN states.}
\label{fig:rel-pred}
\end{figure}
We believe that using discrete alignments, rather than attention-based models ~\cite{bahdanau2014neural} is crucial for AMR parsing. AMR banks are a lot smaller than parallel corpora used in machine translation (MT) and hence it is important to inject a useful inductive bias. 
We constrain our alignments from concepts to words to be injective.
First, it encodes the observation that
concepts are mostly triggered by single words (especially, after re-categorization, Section~\ref{sec:re-cat}). 
Second, it implies that each word corresponds to at most one concept (if any). This encourages competition: alignments are mutually-repulsive. In our example, \textit{obligate} is not lexically similar to the word \textit{must} and may be hard to align. 
However, given that other concepts are easy to predict, 
alignment candidates other than
{\it must} and {\it the} will be immediately ruled out.
We believe that these are the key reasons for why attention-based neural models do not achieve competitive results on AMR~\cite{konstas-EtAl:2017:Long} and why state-of-the-art models rely on aligners. Our goal is to combine best of two worlds: to use alignments (as in state-of-the-art AMR methods) and to induce them while optimizing for the end goal (similarly to the attention component of encoder-decoder models).

Our model consists of three parts: (1) the concept identification model
$P_\theta(\mathbf{c} | \mathbf{a}, \mathbf{w})$; (2) the relation identification model $P_\phi(R | \mathbf{a}, \mathbf{w}, \mathbf{c})$
and (3) the alignment model $Q_\psi(\mathbf{a} | \mathbf{c}, R, \mathbf{w})$.\footnote{$\theta$, $\phi$ and $\psi$ denote all parameters of the models.}
Formally, (1) and (2) together with the uniform prior over alignments $P(\mathbf{a})$ form the generative model of AMR graphs. In contrast,  the alignment model $Q_\psi(\mathbf{a} | \mathbf{c}, R, \mathbf{w})$, as will be explained below, is approximating the intractable posterior $P_{\theta,\phi}(\mathbf{a} |  \mathbf{c}, R, \mathbf{w})$ within that probabilistic model.

In other words, we assume the following model for generating the AMR graph: 
\begin{align}
\nonumber
&P_{\theta,\phi}(\mathbf{c}, R | \mathbf{w})  \!
= \! \sum_{\!\mathbf{a}}{\! P(\mathbf{a}) P_\theta(\mathbf{c} | \mathbf{a}, \mathbf{w}) P_\phi(R | \mathbf{a}, \mathbf{w}, \mathbf{c})} \\
\nonumber
&=\! \sum_{ \mathbf{a}} {\! P(\mathbf{a}) \prod_{i=1}^{m}\! P(c_i | {\mathbf{h}_{a_i}})}
\!\!  \prod_{i,j=1}^{m}{\!\ P(r_{ij} | \mathbf{h}_{a_i},\! \mathbf{c}_i,\! \mathbf{h}_{a_j},\!\mathbf{c}_j)}\\
\nonumber
\end{align}
AMR concepts are assumed to be generated conditional independently relying on the BiLSTM states and surface forms of the aligned words. Similarly, relations are predicted based only on AMR concept embeddings and LSTM states corresponding to words aligned to the involved concepts. Their combined representations are fed into a bi-affine classifier~\cite{Biaffine} (see Figure~\ref{fig:rel-pred}).

The expression involves intractable marginalization over all valid alignments. 
As standard in variational autoencoders, VAEs~\cite{kingma2013auto}, we lower-bound the log-likelihood as 
\begin{align}
\nonumber
&
\log P_{\theta,\phi}(\mathbf{c}, R | \mathbf{w}) 
\\
\nonumber
&
 \geq 
   E_Q [
   \log  P_\theta(\mathbf{c} | \mathbf{a}, \mathbf{w}) P_\phi(R | \mathbf{a}, \mathbf{w}, \mathbf{c}) 
   ] \\
   &
   \label{eq:elbo}
   - D_{KL}( Q_\psi(\mathbf{a} | \mathbf{c}, R, \mathbf{w}) || P(\mathbf{a})),
\end{align}
where $Q_\psi(\mathbf{a} | \mathbf{c}, R, \mathbf{w})$ is the variational posterior (aka the inference network), $E_Q [\ldots]$ refers to the expectation under $Q_\psi(\mathbf{a} | \mathbf{c}, R, \mathbf{w})$ and $D_{KL}$ is the Kullback-Liebler divergence. In VAEs, the lower bound is maximized both with respect to model parameters ($\theta$ and $\phi$ in our case) and the parameters of the inference network ($\psi$). 
Unfortunately, gradient-based optimization with discrete latent variables is challenging. We use a continuous relaxation of our optimization problem, where real-valued vectors $\hat{\mathbf{a}}_i \in \mathbb{R}^n$ (for every concept $i$) approximate discrete alignment variables $a_i$. This relaxation results in low-variance estimates of the gradient using the parameterization trick~\cite{kingma2013auto}, and ensures fast and stable training.
We will describe the model components and the relaxed inference procedure in detail in  \cref{sec:sink,sec:relax}.

Though the estimation procedure requires the use of the relaxation, the learned parser is straightforward to use. Given our assumptions about the alignments, we can independently choose for each word $w_k$ ($k = 1, \ldots, m$) the most probably concept according to $P_\theta(c| \mathbf{h}_{k})$. If the highest scoring option is {\it NULL}, no concept is introduced. The relations could then be predicted relying on $P_\phi(R | \mathbf{a}, \mathbf{w}, \mathbf{c})$. This would have led to generating inconsistent AMR graphs, so instead we search for the highest scoring valid graph (see Section  \ref{sec:post}). Note that the alignment model $Q_\psi$ is not used at test time and only necessary to train accurate concept and relation identification models.

\subsection{Concept identification model}
\label{sect:concept-ident}

The concept identification model chooses a concept $c$ (i.e. a labeled node) conditioned on the aligned word $k$ or decides that no concept should be introduced (i.e. returns \textit{NULL}). Though it can be modeled with a softmax classifier, it would not be effective in handling rare or unseen words. First, we split the decision into estimating the probability of concept category $\tau(c) \in \mathcal{T}$ (e.g. `number', 'frame') and estimating the probability of the specific concept within the chosen category. Second, based on a lemmatizer and training data\footnote{See supplementary materials.}
we prepare one candidate concept $e_k$ for each word $k$ in vocabulary (e.g., it would propose {\it want} if the word is {\it wants}).  Similar to~\newcite{luong2014addressing}, our  model can then either copy the candidate $e_k$  or rely on the softmax over potential concepts of category $\tau$. Formally, the concept prediction model is defined as
\begin{align}
\nonumber
P_\theta(&c | \mathbf{h}_k, w_k) = P(\tau(c) | \mathbf{h}_k, w_k) \times \\
\nonumber
&
\frac{
  [[e_k = c ]] \times \exp( \mathbf{v}_{copy}^T \mathbf{h_k}) + \exp(\mathbf{v}_{c}^T\mathbf{h_k}) 
}{
 Z(\mathbf{h_k}, \theta)
},
\end{align}
where the first multiplicative term is a softmax classifier over categories (including {\it NULL}); $\mathbf{v}_{copy}, \mathbf{v}_{c} \in \mathbb{R}^d$ (for $c \in \mathcal{C}$) are model parameters; $[[\ldots]]$ denotes the indicator function and equals 1 if its argument is true and 0, otherwise; $Z(\mathbf{h}, \theta)$ is the partition function ensuring that the scores sum to 1. 

\subsection{Relation identification model}
\label{sect:rel-mod}

We use the following arc-factored relation identification model:
\begin{equation}
\label{eq:arc-factored}
P_\phi(R | \mathbf{a}, \mathbf{w}, \mathbf{c}) = \prod_{i,j=1}^{m}{\!\ P(r_{ij} | \mathbf{h}_{a_i},\! \mathbf{c}_i,\! \mathbf{h}_{a_j},\!\mathbf{c}_j)} 
\end{equation}
Each term is modeled in exactly the same way: 
\begin{enumerate}
\item for both endpoints, embedding of the concept $c$ is concatenated with the RNN state $\mathbf{h}$; 
\item they are linearly projected to a lower dimension separately through $M_{h}(\mathbf{h}_{a_i} \circ c_i) \in \mathbb{R}^{d_f}$ and $M_{d}(\mathbf{h}_{a_j}  \circ c_j)  \in \mathbb{R}^{d_f}$, where $\circ$ denotes concatenation; 
\item a log-linear 
model with bilinear scores $M_{h}(\mathbf{h}_{a_i} \circ c_i)^T C_r M_{d}(\mathbf{h}_{a_j}  \circ c_j)$, $C_r \in \mathbb{R}^{d_f \times d_f}$
is used to compute the probabilities.
\end{enumerate}

In the above discussion, we assumed that BiLSTM 
encodes a sentence once and the BiLSTM states are then used to predict concepts and relations. In semantic role labeling, the task
closely related to the relation identification stage of AMR parsing, a slight modification of this approach was shown more effective~\cite{zhou2015end,marcheggiani-frolov-titov:2017:CoNLL}. In that previous work, the sentence was encoded by a BiLSTM once per each predicate (i.e. verb) and the encoding was in turn used to identify arguments of that predicate. The only difference across the re-encoding passes was 
a binary flag used as input to the BiLSTM encoder at each word position. The flag was set to 1 for the word corresponding to the predicate and to 0 for all other words. In that way, BiLSTM was encoding the sentence specifically for predicting arguments of a given predicate.
Inspired by this approach, when predicting label $r_{ij}$ for $j \in \{1,\ldots\,m\}$, we input binary flags $\mathbf{p}_1, \ldots \mathbf{p}_n$ to the BiLSTM encoder which are set to $1$ for the word indexed by $a_i$ ($\mathbf{p}_{a_i} = 1$) and to $0$ for other words ($\mathbf{p}_{j} = 0$, for $j \neq a_i$). 
This also means that BiLSTM encoders for predicting relations and concepts end up being distinct. 
We use this multi-pass approach in our experiments.\footnote{Using the vanilla one-pass model from equation~(\ref{eq:arc-factored}) results in 1.4\% drop in Smatch score.}

\subsection{Alignment model}
\label{sec:align-mod}
Recall that the alignment model is only used at training, and hence it can rely both on input (states $\mathbf{h}_1, \ldots, \mathbf{h}_n$) and on the list of concepts $c_1, \ldots, c_m$. 

Formally, we add $(m-n)$
\textit{NULL} concepts to the list.\footnote{After re-categorization (Section~\ref{sec:re-cat}), $m \geq n$ holds for most cases. For exceptions, we append \textit{NULL} to the sentence.} 
Aligning a word to any \textit{NULL}, would correspond to saying that the word is not aligned to any `real' concept. Note that each one-to-one alignment (i.e. permutation) between $n$ such concepts and $n$ words implies a valid  injective alignment of $n$ words to $m$ `real' concepts. This reduction to permutations will come handy when we turn to the Gumbel-Sinkhorn relaxation in the next section. 
Given this reduction, from now on, we will assume that $m = n$. 

As with sentences, we use a BiLSTM model to encode concepts $\mathbf{c}$, where $\mathbf{g}_i \in \mathcal{R}^{d_g}$, $i \in \{1, \ldots, n\}$. We use a globally-normalized alignment model:
\begin{align}
\nonumber
Q_{\psi}( \mathbf{a} | \mathbf{c}, R, \mathbf{w}) =
\frac{
\exp(\sum_{i=1}^{n}{
      \varphi(\mathbf{g}_i, \mathbf{h}_{a_i})
   })
}
{
Z_{\psi} (\mathbf{c}, \mathbf{w}) 
},
\end{align}
where $Z_{\psi} (\mathbf{c}, \mathbf{w})$ is the intractable partition function and the terms $\varphi(\mathbf{g}_i, \mathbf{h}_{a_i})$ score each alignment link according to a bilinear form
\begin{align}
\varphi(\mathbf{g}_i, \mathbf{h}_{a_i}) = \mathbf{g}_i^T B \mathbf{h}_{a_i},
\end{align}
where $B \in \mathbb{R}^{d_g \times d}$ is a parameter matrix.

\subsection{Estimating model with Gumbel-Sinkhorn}\label{sec:sink}

Recall that our learning objective~(\ref{eq:elbo}) involves expectation under the alignment model. The partition function of the alignment model $Z_{\psi} (\mathbf{c}, \mathbf{w})$  is intractable, and it is tricky even to draw samples from the distribution. Luckily, the recently proposed relaxation~\cite{sinkhorn} lets us circumvent this issue. First, note that exact samples from a categorical distribution can be obtained using the perturb-and-max technique~\cite{papandreou2011perturb}. For our alignment model, it would correspond to adding independent noise to the score for every possible alignment and choosing the highest scoring one:
\begin{align}
\label{eq:argmax}
\mathbf{a^{\star}} = \argmax_{\mathbf{a} \in \mathcal{P}}{\sum_{i=1}^{n}{\varphi(\mathbf{g}_i, \mathbf{h}_{a_i})} + \epsilon_{\mathbf{a}}},
\end{align}
where $\mathcal{P}$ is the set of all permutations of $n$ elements,
$\epsilon_{\mathbf{a}}$ is a noise drawn independently for each $\mathbf{a}$ from the fixed Gumbel distribution ($\mathcal{G}(0,1)$). Unfortunately, this is also intractable, as there are $n!$ permutations. Instead, in perturb-and-max
an approximate schema is used where noise is assumed factorizable. In other words, first noisy scores are computed as $\hat{\varphi}(\mathbf{g}_i, \mathbf{h}_{a_i}) = {\varphi}(\mathbf{g}_i, \mathbf{h}_{a_i}) + \epsilon_{i, a_i}$, where $\epsilon_{i, a_i} \sim \mathcal{G}(0,1)$ and an approximate sample is obtained by
$\mathbf{a^{\star}} = \argmax_{\mathbf{a}}{\sum_{i=1}^{n}{\hat{\varphi}(\mathbf{g}_i, \mathbf{h}_{a_i})}},$

Such sampling procedure is still intractable in our case and also non-differentiable.
The main contribution of~\newcite{sinkhorn} is approximating this $\argmax$
with a simple differentiable computation $\hat{\mathbf{a}} = S_t(\Phi, \Sigma)$ which yields an approximate (i.e. relaxed) permutation. We use $\Phi$ and $\Sigma$ to denote the $n \times n$ matrices of alignment scores $\varphi(\mathbf{g}_i, \mathbf{h}_{k})$ and noise variables $\epsilon_{ik}$, respectively.  Instead of returning index $a_i$ for every concept $i$,  it would return a (peaky) distribution over words $\hat{\mathbf{a}}_i$. The peakiness is controlled by the temperature parameter $t$ of Gumbel-Sinkhorn  which balances smoothness (`differentiability') vs. bias of the estimator. 
For further details and the derivation, we refer the reader to the original paper ~\cite{sinkhorn}.

Note that $\Phi$ is a function of  the alignment model $Q_{\psi}$, so we will write $\Phi_\psi$ in what follows. The variational  bound~(\ref{eq:elbo}) can now be approximated as
\begin{align}
\nonumber
&E_{\Sigma \sim \mathcal{G}(0,1)}[\log P_\theta(c | S_t(\Phi_\psi, \Sigma), \mathbf{w}) \\
\nonumber
& \quad + \log P_\phi(R | S_t(\Phi_\psi, \Sigma), \mathbf{w}, \mathbf{c})
]\\
\label{eq:relaxed-elbo} 
  & \quad - D_{KL}(  \frac {\Phi_{\psi} + \Sigma} { t    } || 
  \frac{\Sigma}{t_0}
  )
\end{align}
Following ~\newcite{sinkhorn}, the original KL term from equation~(\ref{eq:elbo}) is approximated by the KL term between two $n \times n$ matrices of i.i.d. Gumbel distributions with different temperature and mean. The parameter $t_0$ is the `prior temperature'. 

Using the Gumbel-Sinkhorn construction unfortunately does not guarantee
that $\sum_i \hat{\mathbf{a}}_{ij} = 1$.
To encourage this equality to hold, and equivalently to discourage overlapping alignments, we add another regularizer to the objective~(\ref{eq:relaxed-elbo}): 
\begin{equation}
\label{eq:sink-reg}
\Omega(\hat{\mathbf{a}},\lambda)= \lambda\sum_j \max(\sum_i (\hat{\mathbf{a}}_{ij})-1,0).
\end{equation}

Our final objective is fully differentiable with respect to all parameters (i.e. $\theta$, $\phi$ and $\psi$) and has low variance as sampling is performed from the fixed non-parameterized distribution, as in standard VAEs.

\subsection{Relaxing concept and relation identification}\label{sec:relax}

One remaining question is how to use the soft input $\hat{\mathbf{a}} = S_t(\Phi_\psi, \Sigma)$ in the concept and relation identification models in equation~(\ref{eq:relaxed-elbo}). In other words, we need to define how we  compute $P_\theta(c | S_t(\Phi_\psi, \Sigma), \mathbf{w})$ and $P_\phi(R | S_t(\Phi_\psi, \Sigma), \mathbf{w}, \mathbf{c})$. 

The standard technique would be to pass to the models expectations under the relaxed variables $\sum_{k=1}^{n}{\hat{\mathbf{a}}_{ik} \mathbf{h}_k}$, instead of the vectors $\mathbf{h}_{a_i}$ \cite{maddison2016concrete,jang2016categorical}. This is what we do for the relation identification model. We use this approach also 
to relax the one-hot encoding of the predicate position ($\mathbf{p}$, see Section~\ref{sect:rel-mod}).

However, the concept prediction model $\log P_\theta(c | S_t(\Phi_\psi, \Sigma), \mathbf{w})$ relies on the pointing mechanism, i.e. directly exploits the words $\mathbf{w}$ rather than relies only on biLSTM states $\mathbf{h}_k$. So instead we treat $\hat{\mathbf{a}}_i$ as a prior in a hierarchical model:
\begin{align}
 \log & P_\theta(c_i | \mathbf{\hat{a}_i}, \mathbf{w}) \nonumber \\ 
 \label{eq:h-model}
& \approx 
 \log \sum_{k=1}^{n}{\hat{\mathbf{a}}_{ik}  P_\theta(c_i | a_i = k, \mathbf{w})  }
\end{align}
As we will show in our experiments, a softer version of the loss 
is even more effective:
\begin{align}
\nonumber
 \log & P_\theta(c_i | \mathbf{\hat{a}_i}, \mathbf{w}) \\
 \label{eq:soft-model}
 &
 \approx 
 {
\log \sum_{k=1}^{n}{  ( \hat{\mathbf{a}}_{ik} P_\theta(c_i | a_i = k, \mathbf{w}) )^{\alpha}  }},
\end{align}
where we set the parameter $\alpha = 0.5$. We believe that using this loss encourages the model to more actively explore the alignment space. Geometrically, the loss surface shaped as a ball in the 0.5-norm space would push the model away from the corners, thus encouraging exploration.

\section{Pre- and post-pocessing}
\label{sect:prepost}

\subsection{Re-Categorization}\label{sec:re-cat}

\begin{figure}[t!]
\centering
\includegraphics[width=0.8\columnwidth]{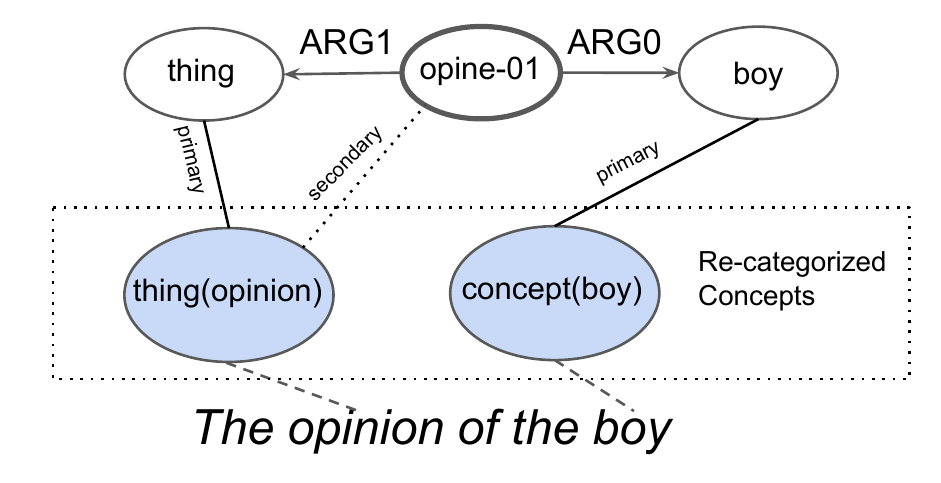}
\vspace{-2ex}
\caption{An example of re-categorized AMR. AMR graph at the top, re-categorized concepts in the middle, and the sentence is at the bottom.}
\label{fig:re-example}
\end{figure}

AMR parsers often rely on a pre-processing stage, where specific subgraphs of AMR are grouped together and assigned to a single node with a new compound category~(e.g., \newcite{werling2015robust,foland-martin:2017:Long,peng2017addressing}); this transformation is reversed at the post-processing stage. Our approach is very similar to the Factored Concept Label system of ~\newcite{wang2017getting}, with one important difference that we unpack our concepts before the relation identification stage, so the relations are predicted between original concepts (all nodes in each group share the same alignment distributions to the RNN states). Intuitively, the goal is to ensure that concepts rarely lexically triggered (e.g., \textit{thing} in Figure~\ref{fig:re-example}) get grouped together with lexically triggered nodes. Such `primary' concepts get encoded in the category of the concept (the set of categories is $\mathcal{\tau}$, see also section~\ref{sect:concept-ident}). In Figure~\ref{fig:re-example}, the re-categorized concept {\it thing(opinion)} is produced from {\it thing} and {\it opine-01}. We use \textit{concept} as the dummy category type.
There are 8 templates in our system which extract re-categorizations for fixed phrases (e.g. {\it thing(opinion)}), and a deterministic system for grouping lexically flexible, but structurally stable sub-graphs (e.g., named entities, {\it have-rel-role-91} and {\it have-org-role-91} concepts).

Details of the re-categorization procedure and other pre-processing are provided in appendix.

\nocite{Pourdamghani2014AligningES,pennington2014glove,manning-EtAl:2014:P14-5}

\subsection{Post-processing}\label{sec:post}
For post-processing, we handle sense-disambiguation, wikification and ensure legitimacy of the produced AMR graph.
For sense disambiguation we pick the most frequent sense for that particular concept (`-01', if unseen). For wikification we again look-up in the training set and default to "-". There is certainly room for improvement in both stages.
Our probability model predicts edges conditional independently and thus cannot guarantee the connectivity of AMR graph, also there are additional constraints which are useful to impose.
We enforce three constraints: (1) specific concepts can have only one neighbor (e.g., `number' and `string'; see appendix for details); (2) each predicate concept can have at most one argument for each relation $r \in  \mathcal{R}$; (3) the graph should be connected.
Constraint (1) is addressed by keeping only the highest scoring neighbor. In order to satisfy the last two constraints we use a simple greedy procedure. First, for each edge, we pick-up the highest scoring relation and edge (possibly {\it NULL}).  If the constraint (2) is violated, we simply keep the highest scoring edge among the duplicates and drop the rest. If the graph is not connected (i.e. constraint (3) is violated), we greedily choose edges linking the connected components until the graph gets connected (MSCG in \newcite{Flanigan_adiscriminative}).

Finally, we need to select a root node. Similarly to relation identification,  for each candidate concept $c_i$, we  concatenate its embedding with the corresponding LSTM state ($\mathbf{h}_{a_i}$) and use these scores in a softmax classifier over all the concepts.

\section{Experiments and Discussion}

\subsection{Data and setting}
\label{sect:setting}
We  primarily focus on the most recent LDC2016E25 (R2) dataset, which consists of 36521, 1368 and 1371 sentences in training, development and testing sets, respectively. The earlier LDC2015E86 (R1) dataset has been used by much of the previous work. It contains 16833 training sentences, and same sentences for development and testing as R2.\footnote{Annotation in R2 has also been slightly revised.} 

 We used the development set to perform model selection and hyperparameter tuning. The hyperparameters, as well as  information about embeddings and pre-processing, are presented in the supplementary materials. 
 
We used Adam~\cite{Kingma2014AdamAM} to optimize the loss (\ref{eq:relaxed-elbo}) and to train the root classifier. 
Our best model is trained fully jointly, and
we do early stopping on the development set scores. Training takes approximately 6 hours on a single GeForce GTX 1080 Ti with Intel Xeon CPU E5-2620 v4.

\nocite{pytorch,nltk}
\begin{table}[t!]
    \begin{center} 
        \begin{tabular}{lll} 
            \hline  Model & Data & Smatch \\ \hline
            JAMR \tiny\cite{jamr-16} & R1& 67.0  \\
            AMREager \tiny\cite{Marco} & R1& 64.0  \\
            CAMR  \tiny\cite{camr_sem} & R1& 66.5  \\
            SEQ2SEQ + 20M  \tiny\cite{konstas-EtAl:2017:Long} & R1&  62.1\\
            Mul-BiLSTM \tiny\cite{foland-martin:2017:Long} & R1&  70.7\\
            Ours & R1&  \bf 73.7\\
            \hline
            Neural-Pointer \tiny\cite{neural-pointer-eval-16} & R2 & 61.9 \\
            ChSeq \tiny\cite{Character} & R2&   64.0\\
            ChSeq + 100K  \tiny\cite{Character} & R2&   71.0\\
            Ours & R2& {\bf74.4} \tiny $ \pm$ 0.16\\
            \hline
        \end{tabular}
    \end{center}
    \caption{\label{table:all_results}Smatch scores on the test set. R2 is LDC2016E25 dataset, and R1 is LDC2015E86 dataset. Statistics on R2 are over 8 runs.}
\end{table}

\subsection{Experiments and discussion}

We start by comparing our parser to previous work (see Table ~\ref{table:all_results}). Our model substantially outperforms all the previous models on both datasets. Specifically, it achieves 74.4\%  Smatch score on LDC2016E25 (R2), which is an improvement of 3.4\% over character seq2seq model relying on silver data~\cite{Character}. For LDC2015E86 (R1), we obtain 73.7\% Smatch score, which is an improvement of 3.0\% over  the previous best model,  multi-BiLSTM parser of~\newcite{foland-martin:2017:Long}.

\begin{table}[t] 
    \begin{center} 
        \begin{tabular}{llll|ll} 
            \hline  Models &  A' &  C'  &   J'  &    Ch'   & Ours\\  
             &17  &16   &16  &17   & \\ \hline
            Dataset &  R1 &R1   &R1   &R2   &R2 \\ \hline
             Smatch &  64 & 63  & 67 & {71} & {\bf74.4}$\pm 0.16 $\\
             \hline
            Unlabeled & 69  & 69&69 &{ 74} &  {\bf77.1}$\pm 0.10 $\\
            No WSD &    65  &64 & 68 & { 72 }&  {\bf75.5}$\pm 0.12$\\
            Reentrancy & 41   & 41  &42 & \bf 52 & {\bf52.3}$\pm 0.43$\\
            Concepts &    83  &80 &83& 82& {\bf 85.9}$\pm 0.11 $\\
            NER &  83  & 75  & 79 &79& \bf{86.0}$\pm 0.46 $\\
            Wiki &   64 &  0  &{75}& 65 &   {\bf75.7}$\pm 0.30 $\\
            Negations&    48  &18 &45 & \bf{62}& {58.4}$\pm 1.32 $\\
            SRL&     56  & 60 & 60 &66& {\bf 69.8}$\pm 0.24 $\\
            \hline
        \end{tabular}
    \end{center}
    \vspace{-1ex}
	\caption{\label{table:evaluation} F1 scores on individual phenomena. A'17 is AMREager, C'16 is CAMR, J'16 is JAMR, Ch'17 is ChSeq+100K. Ours are marked with standard deviation.
    }
    \vspace{-1ex}
\end{table}

In order to disentangle individual phenomena, we use the AMR-evaluation tools~\cite{Marco} and compare to systems which reported these scores (Table \ref{table:evaluation}). We obtain the highest scores on most subtasks. 
The exception is negation detection. However, this is not too surprising as many negations are encoded with morphology, and character models, unlike our word-level model, are able to capture predictive morphological features (e.g., detect prefixes such as ``un-'' or ``im-''). %

\begin{table}[t] 
    \begin{center} 
        \begin{tabular}{lll|ll} 
            \hline  Metric  &  Pre- &  R1 &  Pre-   & R2 \\  
                           &Align    &       &  Align   &mean \\ \hline
            Smatch &           72.8 & 73.7    & 73.5    & {\bf74.4}\\ \hline
            Unlabeled&          75.3 & 76.3   &76.1    &  {\bf77.1}\\
            No WSD &            73.8 & 74.7     & 74.6     &  {\bf75.5}\\
            Reentrancy &         50.2&50.6    &{\bf52.6 }   & 52.3\\
            Concepts &          85.4& 85.5  &85.5      & {\bf 85.9}\\
            NER &           85.3   &84.8  & 85.3      & {\bf86.0}\\
            Wiki &             66.8  & 75.6  &67.8      &   {\bf75.7}\\
            Negations&         56.0 & 57.2    &56.6     & {\bf58.4}\\
            SRL&             68.8 &68.9     &{\bf  70.2  }    & 69.8\\
            \hline
        \end{tabular}
    \end{center}
    \vspace{-2ex}
	\caption{\label{table:ablation} F1 scores of on subtasks. Scores on ablations are averaged over 2 runs.
     The left side results are from LDC2015E86 and right results are from LDC2016E25.}
     \vspace{-1ex}
\end{table}

Now, we turn to ablation tests (see Table~\ref{table:ablation}). First, we would like to see if our latent alignment framework is beneficial. In order to
test this, we create a baseline version of our system  (`pre-align') which relies on the JAMR aligner~\cite{Flanigan_adiscriminative}, rather than 
induces alignments as latent variables. Recall that in our model we used training data and a lemmatizer to produce candidates for the concept prediction model (see Section~\ref{sect:concept-ident}, the copy  function). In order to have a fair comparison, if a concept is not aligned after JAMR, we try to use our copy function to align it. If an alignment is not found, we make the alignment uniform across the unaligned words. In preliminary experiments, we considered alternatives versions (e.g., dropping concepts unaligned by JAMR or dropping concepts unaligned after both JAMR and the matching heuristic), but the chosen strategy was the most effective. 
These scores of pre-align are superior to the results from  \newcite{foland-martin:2017:Long} which also relies on JAMR alignments and uses BiLSTM encoders. There are many potential reasons for this difference in performance. For example,  their relation identification model is different (e.g., single pass, no bi-affine modeling), they used much smaller networks than us,
they use plain JAMR rather than a combination of JAMR and our copy function, they use a different recategorization system.
These results confirm that we started with a strong basic model, and that our variational alignment framework provided further gains in performance.

\begin{figure}[t!]
\centering
\includegraphics[width=0.8\columnwidth]{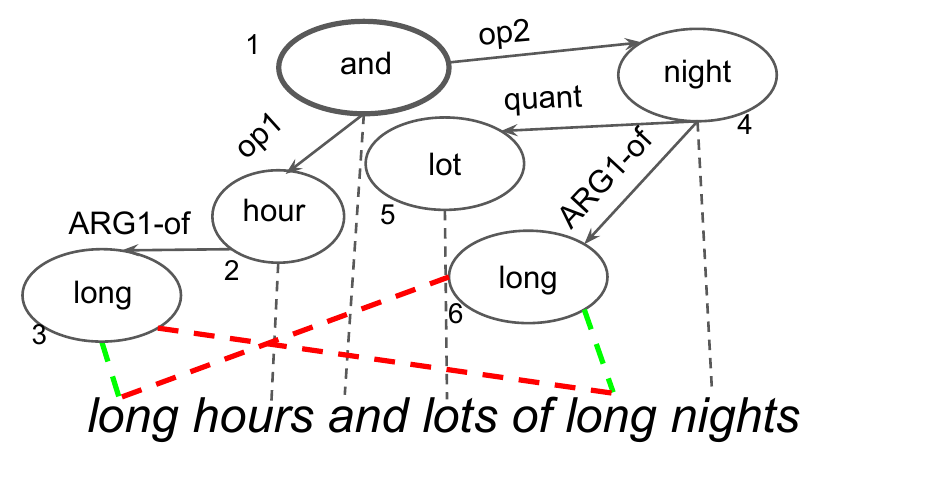}
\vspace{-2ex}
\caption{When modeling concepts alone, the posterior probability of the correct (green) and wrong (red) alignment links will be the same.} 
\label{fig:dup-example}
\end{figure}

\begin{table}[t]
    \begin{center} 
        \begin{tabular}{llll} 
            \hline  Ablation   &   Concepts &    SRL   & Smatch \\\hline
            2 stages     &85.6& 68.9&73.6\\
            2 stages, tune align      & 85.6 & 69.2& 73.9\\
            
            Full model      &\bf85.9& \bf 69.8& {\bf 74.4}\\
            \hline
        \end{tabular}
    \end{center}
    \vspace{-2ex}
	\caption{\label{table:ablation2} Ablation studies: effect of joint modeling (all on R2). Scores on ablations are averaged over 2 runs. The first two models load  the same concept and alignment model before the second stage. 
    }
    \vspace{-2ex}
\end{table}

Now we would like to confirm  that joint training of alignments with both concepts and relations is beneficial. In other words,
we would like to see if alignments need to be induced in such a way as to benefit the relation identification task.
For this ablation
we break the full joint training into two stages.
We start by jointly training the alignment model and the concept identification model. When these are trained, we  optimizing the relation model   but keep the concept identification model and alignment models fixed (`2 stages' in see Table~\ref{table:ablation2}). When compared to our joint model (`full model'),  we observe a substantial drop in Smatch score (-0.8\%). In another version (`2 stages, tune align') we also use two stages but we fine-tune the alignment model on the second stage. This approach appears slightly more accurate but still -0.5\% below the full model.  In both cases, the drop is more substantial for relations (`SRL').
In order to see why relations are potentially useful in learning alignments, consider  Figure~\ref{fig:dup-example}. The example contains duplicate concepts \textit{long}. The concept prediction model factorizes over concepts and does not care which way these duplicates are aligned: correctly (green edges) or not (red edges). Formally, the true posterior under the concept-only model in `2 stages' assigns exactly the same probability to both configurations, and the alignment model $Q_{\psi}$  will be forced to mimic it (even though it relies on an LSTM model of the graph). The spurious ambiguity will have a detrimental effect on the relation identification stage.

\begin{table}[t]
    \begin{center} 
        \begin{tabular}{llll} 
            \hline  Ablation   &   Concepts &    SRL   & Smatch \\\hline
            No Sinkhorn     &85.7& 69.3&73.8\\
            No Sinkhorn reg     &85.6& 69.5&74.2\\ 
            No soft loss      & 85.2 & 69.1& 73.7\\
            Full model     &\bf85.9& \bf69.8& {\bf 74.4}\\
            \hline
        \end{tabular}
    \end{center}
    \vspace{-2ex}
	\caption{\label{table:ablation3} Ablation studies: alignment modeling and relaxation (all on R2). Scores on ablations are averaged over 2 runs. 
    }
    \vspace{-2ex}
\end{table}

It is interesting to see the contribution of other modeling 
decisions we made when modeling and relaxing alignments. First, instead of using Gumbel-Sinkhorn,
which encourages mutually-repulsive alignments, we now use a factorized alignment model. Note that this model (`No Sinkhorn' in
Table~\ref{table:ablation3}) 
still relies on (relaxed) discrete alignments (using Gumbel softmax) but does not constrain the alignments to be injective.
A substantial drop in performance indicates that the prior knowledge about  the nature of alignments appears beneficial.
Second, we remove the additional regularizer for Gumbel-Sinkhorn approximation (equation~(\ref{eq:sink-reg})). The performance drop
in Smatch score (`No Sinkhorn reg') is only moderate.
Finally, we show that using the simple hierarchical relaxation (equation~(\ref{eq:h-model})) rather than our softer version of the loss (equation~(\ref{eq:soft-model}))
results in a substantial drop in performance (`No soft loss', -0.7\% Smatch). We hypothesize that the softer relaxation favors exploration of alignments and helps to discover better configurations.

\section{Additional Related Work}
Alignment performance has been previously identified as a potential bottleneck affecting AMR parsing~\cite{Marco,foland-martin:2017:Long}. Some recent work  has focused on building aligners specifically for training their parsers~\cite{werling2015robust,wang2017getting}. 
However, those aligners are trained independently of concept and relation identification and only used at  pre-processing. 

Treating alignment as discrete variables has been successful in some sequence transduction tasks with neural models \cite{Noisy_Channel,seg_to_seg}. Our work is similar in that we also train discrete alignments jointly but the tasks, the inference framework and the decoders are very different. 

The discrete alignment modeling framework has been developed in the context of traditional (i.e. non-neural) statistical machine translation~\cite{IBM}. 
Such translation models have also been
successfully applied to semantic parsing tasks~(e.g., \cite{andreas2013semantic}), where they rivaled specialized semantic parsers from that period. However, they are considerably less accurate than current state-of-the-art parsers applied to the same datasets (e.g., \cite{dong2016language}).

For AMR parsing, another way to avoid using pre-trained aligners is to use seq2seq models \cite{konstas-EtAl:2017:Long,Character}. In particular, \newcite{Character} used character level seq2seq model and achieved the previous state-of-the-art result. However, their model is very data demanding as they needed to train it on additional 100K  sentences parsed by other parsers. 
This may be due to two reasons. First, seq2seq models are often not as strong on smaller datasets. Second, recurrent decoders may struggle with predicting the linearized AMRs, as many statistical dependencies are highly non-local.

\section{Conclusions}
We introduced a neural AMR parser trained by jointly modeling alignments, concepts and relations. We make such joint modeling computationally feasible by using the variational auto-encoding framework and continuous relaxations.  
The parser achieves state-of-the-art results and ablation tests show that joint modeling is indeed beneficial. 

We believe that the proposed approach may be extended to other parsing tasks where alignments are latent (e.g., parsing to logical form~\cite{liang2016learning}). Another promising direction is integrating character seq2seq to substitute the copy function. This should also improve the handling of negation and rare words.
Though our parsing model does not use any linearization of the graph, we relied on LSTMs and somewhat arbitrary linearization (depth-first traversal) 
to encode the AMR graph in our alignment model. A better alternative would be to use graph convolutional networks~\cite{marcheggiani2017encoding,kipf2017semi}: neighborhoods in the graph are likely to be more informative for predicting alignments  than  the neighborhoods in the graph traversal.

\section*{Acknowledgments}
We thank Marco Damonte, Shay Cohen, Diego Marcheggiani and Wilker Aziz for helpful discussions as well as anonymous reviewers for their 
suggestions. The project was supported by the
European Research Council (ERC StG BroadSem
678254) and the Dutch National Science Foundation
(NWO VIDI 639.022.518).
\bibliography{acl2018}
\bibliographystyle{acl_natbib}
\onecolumn 
\section*{\LARGE\centering{Supplementary Material }}
\vspace{4cm}
\section{Matching algorithm for copying concepts}

Only frequent concepts $c$ (frequency at least 10 for R2 and 5 for R1) can be generated without the copying mechanism (i.e. have their own vector $\mathbf{v}_c$ associated with them). Both frequent and infrequent ones are processed with coping, using candidates produced by the algorithm below and the matching rule in Table~\ref{table:str}.

{
\linespread{1}
\begin{algorithm}
\SetAlgoLined
\SetKwInOut{Input}{Input}\SetKwInOut{Output}{Output}
\Input{ $\{ \mathbf{w}^l,\mathbf{c}^l\}^{L}_{l=1}$}
\Output{ D  copy dictionary}
Counter $\leftarrow \emptyset$ \\

 \For{$l = 1$ to $L$}{
  \For{all pairs $c_i^l$ and $w_j^l$}{ 
 \If{ match($c_i^l$, $w_j^l$)}{
Increment Counter[$w_j^l$][$c_i^l$] 
 }
 }}
 D $\leftarrow $ default Stanford lemmatizer\\
 \For{w $\leftarrow $ Counter}{
 D[w] $\leftarrow \operatorname*{argmax}_{c}  $ Counter[w][c]
 }
 \Return D
 \caption{Copy function construction} \label{construct}
\end{algorithm}
}

\begin{table*}[ht!]
    \begin{center} 
        \begin{tabular}{ll} 
            \hline  Rules &   Matching Criteria \\  \hline
             Verbalization Match & exact match frame in "verbalization-list-v1.06.txt"\\
            PropBank  Match  & exact match frame in PropBank frame files\\
             Suffix Removal Match  &word with suffix (``-ed'', ``-ly'',``-ing'') removed is identical to concept lemma\\
             Edit-distance Match  &edit distance smaller than 50\% of the length  \\
            \hline
        \end{tabular}
    \end{center}
	\caption{\label{table:str} Matching rules for Algorithm 1}
\end{table*}

\section{Re-categorization details}
Re-categorization is handled with rules listed in Table 2. They are triggered if a given primary concept (`primary') appears adjacent to edges labeled with relations given in column `rel'. The assigned category is shown in column `re-categorized'. The rules yield 32 categories when applied to the training set.

\begin{table*}[t!] \small 
    \begin{center} 
        \begin{tabular}{llll} 
            \hline  primary &  rel    &  re-categorized \\  \hline
            person &  ARG0-of/ARG1-of & person([second])\\
            thing &  ARG0-of/ARG1-of/ARG2-of & thing([second])\\
            most & degree-of &   most([second])\\
            -quantity & unit  &    primary([second])\\
            date-entity & weekday/dayperiod/season  &   date-entity([second])\\
            monetary-quantity & unit/ARG2-of/ARG1-of/quant  &   monetary-quantity([second])\\
            temporal-quantity & unit/ARG3-of  &    temporal-quantity([second])\\
            \hline
        \end{tabular}
    \end{center}
	\caption{\label{table:templates} Templates for re-categorization.}
\end{table*}

There are also rules of another type shown in Table 3 below. The templates and examples are in column `original', the resulting concepts are in column `re-categorized'. These rules yield 109 additional types when applied to the training set.

\begin{table*}[ht!] \small 
    \begin{center} 
        \begin{tabular}{ll} 
            \hline  original    &  re-categorized \\  \hline
            
            \begin{lstlisting}
(c / type
    :name (n / name
        :op1 `n1'
        ...
        :opx `nx')
\end{lstlisting}& (B-Ner\_type(n1),...,Ner\_type(nx))\\ 
            \begin{lstlisting}
(c / city
    :name (n / name
        :op1 `New'
        :op2 `York')
\end{lstlisting}& B-Ner\_city(New),Ner\_city(York)\\ 
\hline \\
            \begin{lstlisting}
(p / type
    :ARG0-of (h / have-x-role-91
        :ARG2 (p / role)
\end{lstlisting}& have-x-role\_type(role)\\ 
            \begin{lstlisting}
(p / person
    :ARG0-of (h / have-org-role-91
        :ARG2 (p / premier)
\end{lstlisting}& {have-org-role\_person(premier)}\\ 
\hline\\
            \begin{lstlisting}
(o1 / x-entity
    :x constant)
\end{lstlisting}& x-entity(constant)\\ 
            \begin{lstlisting}
(o1 / ordinal-entity
    :value 1)
\end{lstlisting}&  ordinal-entity(1)\\ 
\hline
            \hline
        \end{tabular}
    \end{center}
	\caption{\label{table:det} Extra rules for re-categorization.  }
\end{table*}

\section{Additional pre-processing}

Besides constructing re-categorized AMR concepts, we perform additional preprocessing. We start with tokenized dataset of \citet{Pourdamghani2014AligningES}. We take all dashed AMR concepts (e.g, {\it make-up} and {\it more-than}) and concatenate the corresponding spans (based on statistics from training set and PropBank frame files). We also combine spans of words corresponding to a single number. For relation identification, we normalize relations to one canonical direction (e.g. arg0, time-of).
For named entity recognition, and lemmatization, we use Stanford CoreNLP toolkit \cite{manning-EtAl:2014:P14-5}. For pre-trained embedding, we used Glove (300 dimensional embeddings)~\cite{pennington2014glove}.  

\section{Model parameters and optimization details}
We selected hyper-parameters based on the best performance on the development set. For all the ablation tests, the hyper parameters are fixed. We used 2 different BiLSTM encoders of the same  hyper-parameters to encode sentence for concept identification and alignment prediction, another BiLSTM to encode AMR concept sequence for alignment, and finally 2 different BiLSTM of the same hyper-parameters to encode sentence for relation identification and root identification.  There are 5 BiLSTM encoders in total. Hyper parameters for the model are summarized in Table~\ref{table:parameter}, and optimization parameters are summarized in Table~\ref{table:opt}. 
\begin{table}[t!]\small
    \begin{center} 
        \begin{tabular}{ll} 
            \hline  Model components &  Hyper-parameters \\  \hline
          Glove Embeddings & 300\\
            Lemma Embeddings & 200\\
            POS Embeddings & 32\\
            NER Embeddings & 16\\
            Category Embeddings & 32\\ \hline 
           Concept/Alignment & 1 layer 548 input \\
          Sentence BiLSTM & 256 hidden (each direction)\\ \hline 
            AMR Categories $\mathcal{T}$& 32\\
            AMR Lemmas $\mathcal{C}$  & 506\\
           AMR NER types & 109\\ \hline 
           Alignment   & 1 layer 232 input \\
           AMR BiLSTM  & 100 hidden (each direction)\\ \hline 
            $B$ bilinear align  & 200 $\times$ 512\\ \hline 
            Relation map dimensionality $d_g$  & 200 \\ \hline 
           Relation/Root &2  layers 549 input \tiny(predicate position)  \\
            Sentence BiLSTM&  256 hidden (each direction)\\ \hline 
            $d_f$ relation vector& 200\\
           $v_{c},v_{copy}$ lemma vector & 512\\
           $v_{root}$ root vector & 200\\
           \hline
           Sinkhorn temperature & 1 \\
           Sinkhorn prior temperature & 5 \\
Sinkhorn steps l for full joint training & 10 \\
Sinkhorn steps l for two stages training & 5 \\
         $\lambda$  & 10 \\
 Dropout & .2 \\
            \hline
        \end{tabular}
    \end{center}
	\caption{\label{table:parameter} Model hyper-parameters}
\end{table}
\begin{table}[ht!]\small
    \begin{center} 
        \begin{tabular}{ll} 
            \hline  Optimizer Parameters &  Values\\  \hline
           Batch size for single stage & 64\\
 Maximum Epochs & 30 \\
           Batch size for first stage & 512\\
           Batch size for second stage & 64\\
 Maximum Epochs for both stages  & 30 \\\hline
         Learning Rate & 1e-4\\
            Adam betas & (0.9, 0.999) \\
          Adam eps & 1e-8 \\
         Weight decay  & 1e-5 \\
            \hline
        \end{tabular}
    \end{center}
	\caption{\label{table:opt} Optimization parameters for full joint training and two stages training.}
\end{table}
\end{document}